\pdfoutput=1

\documentclass[11pt]{article}

\usepackage[]{ACL2023}

\usepackage{times}
\usepackage{latexsym}

\usepackage[T5]{fontenc}

\usepackage[utf8]{inputenc}

\usepackage{microtype}

\usepackage{inconsolata}

\usepackage{textcomp}
\usepackage{amsmath,amssymb,amsfonts}
\usepackage{graphicx}
\usepackage{subcaption}
\usepackage{xcolor}
\usepackage{booktabs}
\usepackage{multirow}
\usepackage{longtable}

\title{A Large-Scale Benchmark for Vietnamese Sentence Paraphrases}

\author{Sang Quang Nguyen\textsuperscript{1, 2}, Kiet Van Nguyen\textsuperscript{1, 2} \\
\textsuperscript{1}University of Information Technology, Ho Chi Minh City, Vietnam\\
\textsuperscript{2}Vietnam National University, Ho Chi Minh City, Vietnam \\
\texttt{sangnq.19@grad.uit.edu.vn} \\
\texttt{kietnv@uit.edu.vn}}

\begin{document}
\maketitle
\begin{abstract}

This paper presents ViSP, a high-quality Vietnamese dataset for sentence paraphrasing, consisting of 1.2M original–paraphrase pairs collected from various domains. The dataset was constructed using a hybrid approach that combines automatic paraphrase generation with manual evaluation to ensure high quality. We conducted experiments using methods such as back-translation, EDA, and baseline models like BART and T5, as well as large language models (LLMs), including GPT-4o, Gemini-1.5, Aya, Qwen-2.5, and Meta-Llama-3.1 variants. To the best of our knowledge, this is the first large-scale study on Vietnamese paraphrasing. We hope that our dataset and findings will serve as a valuable foundation for future research and applications in Vietnamese paraphrase tasks. The dataset is available for research purposes at \url{https://github.com/ngwgsang/ViSP}.

\end{abstract}

\section{Introduction}

Sentences or phrases that express the same idea but use different words are called paraphrases \cite{whatisparaphrase}. Paraphrase helps create a richer amount of data, but still retains the main meaning of the sentence used.


Paraphrases generation is crucial for various tasks such as: In question answering \cite{bernhard2008answering, dong2017learningparaphrasequestionanswering, gan-ng-2019-improving}, by generating paraphrases of the retrieved answers, QA systems can provide more comprehensive and nuanced responses; In information retrieval \cite{wallis1993information, zukerman2002experiments}, paraphrasing can help search engines find relevant documents even if the user's query doesn't match the exact wording of the documents; Machine translation \cite{callison2006improved, russo2005paraphrase} , paraphrasing techniques can enhance translation accuracy by generating more natural and semantically equivalent translations and chat bot \cite{marceau2022quickstartingdialogsystems}, paraphrasing enables chat bot to respond more flexibly and naturally to user queries, adapting to variations in phrasing.

Although Vietnamese is widely spoken languages, Vietnamese is referred to as a low-resource language in NLP. Most previous work in paraphrase generation has focused mainly on English, such as 
MS-COCO \cite{mscoco}, 
PAWS \cite{alzantot2018generatingnaturallanguageadversarial},
QQP~\footnote{\url{https://quoradata.quora.com/First-Quora-Dataset-Release-Question-Pairs}},
ParaSCI \cite{ParaSCI}.
Although there are multilingual datasets such as TaPaCo \cite{scherrer-2020-tapaco}, the number of Vietnamese sentence pairs is only 962, the number of sentence pairs is too small and because they are translated from English, the meaning will not be fluent. Some other works related to paraphrasing, such as ViQP \cite{nguyen2023viqp}, have the limitation that their scope is only in questions, and questions in Vietnamese have a completely different structure than normal sentences.

In this paper, two our main contributions are described as follows:
\begin{enumerate}
\item \textbf{The creation of ViSP, the first large-scale dataset for Vietnamese sentence paraphrasing.} 
    We developed a dataset containing over 1.2 million pairs of Vietnamese sentences across diverse topics. Each original sentence is accompanied by multiple paraphrases, all manually verified by a team of annotators to ensure high quality and accuracy.

    \item \textbf{Comprehensive exploration of Vietnamese sentence paraphrasing.} 
We evaluated baseline models and compared their performance with traditional methods, such as rule-based approaches and back translation, as well as human performance. This analysis highlights the relative strengths and limitations of automated paraphrase generation for Vietnamese.
\end{enumerate}

We hope ViSP together with our empirical study can serve as a starting point for future Vietnamese paraphrase research and applications.





\section{Dataset Creation}

\begin{figure}[h]
    \centering
    \includegraphics[width=0.48\textwidth]{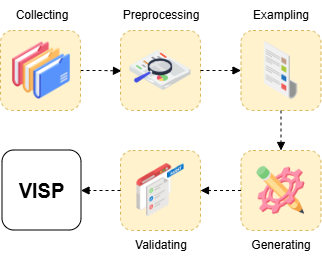}
    \caption{The overview process of creating our dataset ViSP.}
    \label{fig:visp_process}
\end{figure}

In this section, we introduce the process of constructing the ViSP dataset (see  Figure \ref{fig:visp_process}), which includes Collecting, Preprocessing, Exampling, Generating and Validating.

\subsection{Data Collection}
 We collect sentences from publicly available resources that contain original Vietnamese documents, including the UIT-ViQuAD \cite{vannguyen2020vietnamesedatasetevaluatingmachine}, UIT-ViNewsQA \cite{10.1145/3527631}, ALQAC \cite{nguyen2023summary} and ViNLI \cite{huynh-etal-2022-vinli} datasets. These datasets provide a diverse range of data sourced from Vietnamese news articles and Wikipedia, offering valuable material for sentence paraphrasing task, respectively.

After collecting data from the available datasets, we proceed to extract sentences from context segments of the above data sources using underthesea, a Vietnamese NLP toolkit\footnote{\url{https://github.com/undertheseanlp/underthesea}}.

\subsection{Preprocessing}
\label{phase_processing}

First, we manually filtered the sentences to remove those that were incorrect, unsuitable for Vietnamese language norms, or contained offensive language. 

Next, we classified the sentences based on their topic using the Gemini \cite{team2023gemini}. The model categorized sentences into various labels, including health, society, lifestyle, science, culture, computer, law, sports, business and other. This step allowed us to organize the sentences by their subject matter, offering a comprehensive overview of different domains within the Vietnamese language context.








\subsection{Exampling}
\label{phase_exampling}

To evaluate the generative performance of the Gemini model, we divided the team into two groups: the generation group  \(\{ H_1, H_2 \} \)  and the evaluation group \(\{ H_3, H_4, H_5, H_6, H_7 \} \). We randomly selected 350 sentences, consisting of 300 for testing and 50 for generate the Few-shot prompt, referred to as the few-shot corpus. 
The annotators in the evaluation group were tasked with manually generating paraphrases for the selected sentences, followed by cross-validation of the paraphrases among the evaluators. 
The generation group individually crafted paraphrases manually, providing a direct comparison against the AI group \(\{ G_1, G_2, G_3 \} \) . We split the dataset into 6 rounds \(\{ R1, R2, R3, R4, R5, R6 \} \), each consisting of 50 sentences. The Few-shot prompts were randomly selected from 10 out of the 50 samples in the Few-shot corpus, which had been created by the evaluation group.

\begin{table}[h]
\centering
\scalebox{0.9}{%
\begin{tabular}{lcccccc}
\toprule
\textbf{} & \textbf{R1} & \textbf{R2} & \textbf{R3} & \textbf{R4} & \textbf{R5} & \textbf{R6} \\
\midrule
\textit{\( G_1 \)} & \textbf{73.10} & 68.30 & \textbf{70.78} & 69.56 & \textbf{70.87} & 68.18 \\
\textit{\( G_2 \)} & 69.51 & 68.74 & 68.98 & 69.45 & 70.44 & 65.55 \\
\textit{\( G_3 \)} & 70.28 & 66.93 & 68.45 & 69.50 & 69.16 & 65.09 \\
\textit{\( H_1 \)} & 72.11 & 66.79 & 70.64 & \textbf{70.25} & 70.56 & 67.13 \\
\textit{\( H_2 \)} & 71.08 & \textbf{69.13} & 70.75 & 68.30 & 69.55 & \textbf{69.48} \\
\bottomrule
\end{tabular}
}
\caption{Compare Gemini with Few-shot examples performance and human performance across six rounds on the BLEU-4.}
\label{tab:fewshot_eval_bleu4}
\end{table}

Table~\ref{tab:fewshot_eval_bleu4} demonstrate that Gemini significantly outperforms human efforts in paraphrase generation across multiple rounds. Specifically, the model achieved a win rate of 83.33\% against \( H_1 \) and 66.67\% against \( H_2 \). These results underscore the effectiveness of AI in replacing manual paraphrase generation, offering both cost savings and greater coverage. 


\subsection{Data Generation}

\begin{table}[h]
\centering
\begin{tabular}{p{7.2cm}}
\toprule
\textbf{Input} \\
\midrule
\textbf{s:} \small Berlin trở thành địa điểm thành phố được viếng thăm nhiều thứ ba tại châu Âu. (\textit{English: Berlin becomes the third most visited city in Europe.}) \\
\small \textbf{k:} 2  \\ 
\midrule
\textbf{Output} \\
\midrule
\small \textbf{\( p_1 \):} Berlin \textcolor{blue}{là thành phố được du khách} viếng thăm nhiều thứ ba tại châu Âu. (\textit{English: Berlin is the third most visited city in Europe \textcolor{blue}{by tourists}.} )

\small \textbf{\( p_2 \):} \textcolor{blue}{Xếp thứ ba về số lượng du khách} viếng thăm tại châu Âu là thành phố Berlin.
(\textit{English:\textcolor{blue}{ Ranked third in terms of number of visitors} in Europe is the city of Berlin.})
\\
\bottomrule
\end{tabular}
\caption{Example of input and output of sentence paraphrase task.}
\label{tab:generating_input_output}
\end{table}

We used the highest-performing prompt from section~\ref{phase_exampling} to generate paraphrases from the cleaned and labeled dataset of original sentences from section~\ref{phase_processing}. The paraphrase generation task can be formalized as follows. For each input, consisting of an original sentence \textit{s}, the number \textit{k} of paraphrases to be generated, and the chosen Few-shot prompt \textit{f}, the model \textit{M} generates a set of paraphrases using the formula \ref{eq:task_visp}:

\begin{equation} 
M_f(s, k) = \{ p_1, p_2, \ldots, p_k \}
\label{eq:task_visp}
\end{equation}

In this setup, the task is to generate \textit{k} paraphrases \( \{ p_1, p_2, \ldots, p_k \} \) that convey the original meaning while varying the structure and wording of the sentence \textit{s}.






\subsection{Data Validation}
Automatic evaluation of the generation results from large language models (LLMs) can be easily achieved when ground truths from existing datasets are available \cite{zhu2023chatgptreproducehumangeneratedlabels}. However, open-ended data like paraphrasing or translation, human validation is necessary \cite{long2024llmsdrivensyntheticdatageneration}. A straightforward idea is to provide some generated samples to human experts, who will then determine whether they are correct.

We established a review process involving seven annotators to ensure the quality of the paraphrased sentences generated by Gemini \cite{team2023gemini}. Each original-paraphrase sentence pair was evaluated by three annotators, corresponding to three votes. Annotators assessed each pair as True or False. A pair was considered valid if it received at least two True votes out of three. Sentence pairs were marked as False if their meaning was not preserved after paraphrasing or if they contained grammatical or spelling errors, based on a checklist (See Appendix~\ref{append:paraphrase-checklist}). A pair that received two or more False votes were removed from the dataset.

\begin{table}[h]
\centering
\begin{tabular}{p{7.2cm}}
\toprule
\textbf{Original} \\
\midrule
\small SpaceX đang thử nghiệm các nguyên mẫu tàu tại cơ sở của họ ở nam Texas, tuy nhiên cả 4 phiên bản bay thử gần đây đều \textcolor{red}{kết thúc bằng vụ nổ}. 

\small (\textit{English: SpaceX is testing the prototypes of the spacecraft at their facility in southern Texas; however, all four recent test flights \textcolor{red}{have ended in explosions}.})
\\
\midrule
\textbf{Paraphrase} \\ 
\midrule
\small Các mẫu tàu đang được thử nghiệm tại cơ sở SpaceX ở phía nam Texas, nhưng cả 4 phiên bản thử nghiệm bay gần đây đã \textcolor{red}{không thành công}. 

\small (\textit{English: The spacecraft prototypes are being tested at SpaceX's facility in southern Texas, but all four recent test flights \textcolor{red}{have failed}.})
\\
\bottomrule
\end{tabular}
\caption{Example of an incorrect paraphrase pair violating the \MakeUppercase{Semantic Equivalence} constraint.}
\label{tab:paraphrase_err_example}
\end{table}

Table~\ref{tab:paraphrase_err_example} presents an example of SEMANTIC EQUIVALENCE error for the generated paraphrases. Sentences with errors were removed from the dataset to ensure high-quality standards. Across the entire dataset, the average error rate was 4.49\% (See Appendix \ref{append:paraphrase-checklist}).

\subsection{Dataset Analysis}

\subsubsection{Overall Statistics}

The statistics of the training, validation and test sets of the ViSP dataset are described in Table~\ref{tab:visp_overall}. 
In the table, we present number of original, the average number of paraphrases per original, the average lengths of original and paraphrased sentences, as well as the vocabulary sizes for both original and paraphrased sentences across all sets. 

\begin{table}[ht]
\centering
\scalebox{0.78}{
\begin{tabular}{lccc}
\toprule
\textbf{Statistics} & \textbf{Train} & \textbf{Val} & \textbf{Test} \\
\midrule
Sentence Pair$^\dagger$                & 406,308  & 391,044  & 380,590 \\
Original                     & 33,030   & 6,929    & 6,963 \\
Avg. paraphrase per original    & 2.97     & 6.91     & 6.80 \\
Avg. original length          & 21.90    & 21.47    & 21.53 \\
Avg. paraphrase length        & 22.95    & 23.36    & 23.35 \\
Original vocab                & 42,135   & 15,826   & 15,952 \\
Paraphrase vocab              & 45,460   & 20,277   & 20,248 \\
\bottomrule
\end{tabular}
}
\caption{Statistics of the training, validation, and test sets of the ViSP dataset. $\dagger$ denotes that total number of paraphrase pairs generated from all possible combinations.}
\label{tab:visp_overall}
\end{table}

\subsubsection{Data Faithfulness and Diversity}

We evaluate the dataset using BLEU \cite{papineni2002bleu} and ROUGE \cite{lin-2004-rouge} to measure semantic similarity between generated paraphrases and original sentences by comparing n-grams. As shown in Table \ref{tab:unified_evaluation}, BLEU-4 scores for the Train, Val, and Test sets are 63.66, 67.24, and 66.83, while ROUGE-2 scores are 72.4, 73.32 and 72.99, indicating strong semantic alignment across all subsets. 

To assess paraphrase diversity, we use DIST-1 and DIST-2\cite{li-etal-2016-diversity}, which measure unique unigrams and bigrams, as well as Entropy-based metrics ENT-4, which capture the distributional richness of generated paraphrases, and Jaccard, which gauges lexical overlap. The DIST-1 scores are 94.94, 95.56, and 95.42, and the DIST-2 scores are 94.74, 95.01, and 94.96 for the Train, Validation, and Test sets, respectively. Additionally, the ENT-4 scores are 5.71, 6.52, and 6.51, while the Jaccard scores are 53.85, 51.61, and 51.39 across the same sets. The consistently high Distinct and ENT-4 values, accompanied by the lower Jaccard on the validation and test sets, suggest that the paraphrases exhibit a diverse lexical distribution, minimizing redundancy while maintaining coherence. The slightly higher diversity metrics in these sets also indicate that the paraphrases are more varied, improving evaluation robustness by ensuring broader linguistic diversity.

\begin{table}[ht]
\centering
\scalebox{0.9}{
\begin{tabular}{lcccccc}
\toprule
\textbf{Type}   & \textbf{Metric}   & \textbf{Train} & \textbf{Val} & \textbf{Test} \\
\midrule
\multirow{2}{*}{\textbf{Semantic}}  
& BLEU-4        & 63.66    & 67.24     & 66.83 \\
& ROUGE-2       & 72.40    & 73.32     & 72.99 \\
\midrule
\multirow{4}{*}{\textbf{Diversity}}
& DIST-1        & 94.94    & 95.56    & 95.42 \\
& DIST-2        & 94.74    & 95.01    & 94.96 \\
& ENT-4         & 5.71     & 6.52     & 6.51 \\
& Jaccard       & 53.85    & 51.61    & 51.39 \\
\midrule
\multirow{4}{*}{\textbf{Human Eval}}
& INF  & 4.74     & 4.73      & 4.78  \\
& REL  & 4.64     & 4.50      & 4.71  \\
& FLU  & 4.86     & 4.83      & 4.80  \\
& COH  & 4.86     & 4.90      & 4.89  \\
\bottomrule
\end{tabular}
}
\caption{Evaluation of semantic faithfulness, diversity, and human evaluation metrics on the Train, Validation, and Test sets.}
\label{tab:unified_evaluation}
\end{table}

Additionally, we conduct a manual evaluation by human experts on 200 randomly selected samples from each of the train, validation, and test sets. Human evaluators assess the paraphrases using a 5-point scale across four key dimensions, based on \cite{grusky2018newsroom}: INF, REL, FLU and COH (see Appendix~\ref{append:human-eval}). Before the evaluation, we measured inter-annotator agreement using Fleiss' Kappa \cite{fleiss1971measuring} for the task of rating paraphrase sentence pairs with five labels corresponding to the 5-point scale (ratings from 1 to 5). The Fleiss' Kappa values for four human metrics was 0.7252, 0.7144, 0.7634, and 0.7481, respectively. According to the interpretation guidelines by \cite{landis1977measurement}, these Kappa values indicate substantial agreement among the annotators. As shown in Table~\ref{tab:unified_evaluation}, the scores were rated quite well, ranging from 4.71 to 4.89.





\subsubsection{Topic Based Analysis}

In Table \ref{tab:topic_distribution}, \textit{Health} and \textit{Society} are the most common topics, making up about 33\% and 19\% of the total dataset, respectively. This disparity occurs because the dataset originates from UIT-ViNewsQA and UIT-ViQuAD (see Appendix~\ref{append:dataset-detail}), which primarily focus on these two topics. The other topics are more evenly spread, each covering around 3\% to 6\% of the data.

\begin{table}[ht]
\centering
\scalebox{1}{
\begin{tabular}{lcccc}
\toprule
\textbf{Topic} & \textbf{Train} & \textbf{Val} & \textbf{Test} & \textbf{All} \\
\midrule
Health              & 11,381  & 2,443   & 2,367   & 16,193 \\
Society    & 7,088   & 1,183   & 1,222   & 9,495  \\
Culture             & 2,189   & 425     & 407     & 3,023  \\
Computer            & 1,669   & 494     & 475     & 2,640  \\
World               & 3,192   & 399     & 415     & 4,008  \\
Sports              & 1,401   & 387     & 405     & 2,195  \\
Science             & 2,001   & 593     & 616     & 3,212  \\
Lifestyle           & 1,947   & 522     & 507     & 2,978  \\
Law                 & 1,804   & 327     & 377     & 2,510  \\
Business            & 1,045   & 342     & 330     & 1,719  \\
\bottomrule
\end{tabular}
}
\caption{Distribution of topics across the Train, Validation, Test, and All sets in the dataset, statistics based on the number of original sentences. For examples of each topic of sentence, see the Appendix Table~ \ref{tab:visp_example_group_by_topic}.}
\label{tab:topic_distribution}
\end{table}

\begin{figure*}[h]
    \centering
    \begin{subfigure}[b]{0.19\textwidth}
        \centering
        \includegraphics[width=\textwidth]{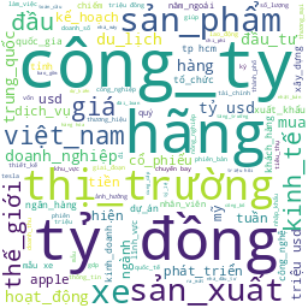}
        \caption{Business}
        \label{fig:sub1}
    \end{subfigure}
    \hfill
    \begin{subfigure}[b]{0.19\textwidth}
        \centering
        \includegraphics[width=\textwidth]{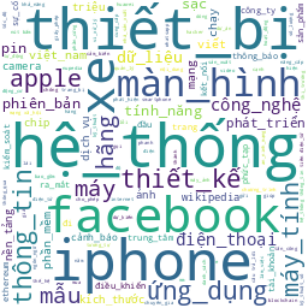}
        \caption{Computer}
        \label{fig:sub2}
    \end{subfigure}
    \hfill
    \begin{subfigure}[b]{0.19\textwidth}
        \centering
        \includegraphics[width=\textwidth]{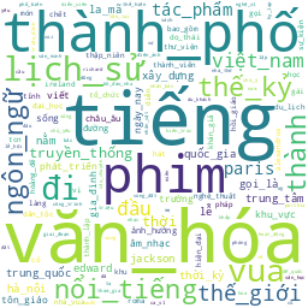}
        \caption{Culture}
        \label{fig:sub3}
    \end{subfigure}
    \hfill
    \begin{subfigure}[b]{0.19\textwidth}
        \centering
        \includegraphics[width=\textwidth]{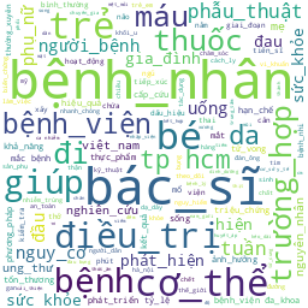}
        \caption{Health}
        \label{fig:sub4}
    \end{subfigure}
    \begin{subfigure}[b]{0.19\textwidth}
        \centering
        \includegraphics[width=\textwidth]{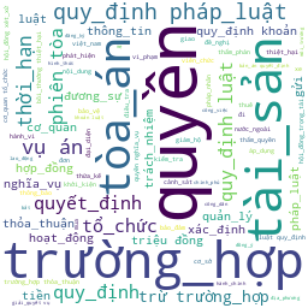}
        \caption{Law}
        \label{fig:sub5}
    \end{subfigure}
    
    \vskip\baselineskip 

    \begin{subfigure}[b]{0.19\textwidth}
        \centering
        \includegraphics[width=\textwidth]{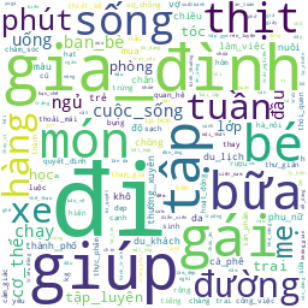}
        \caption{Lifestyle}
        \label{fig:sub6}
    \end{subfigure}
    \hfill
    \begin{subfigure}[b]{0.19\textwidth}
        \centering
        \includegraphics[width=\textwidth]{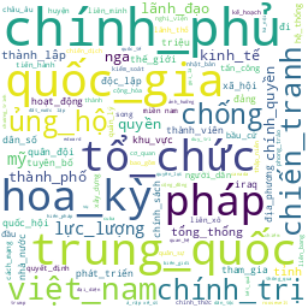}
        \caption{Society}
        \label{fig:sub7}
    \end{subfigure}
    \hfill
    \begin{subfigure}[b]{0.19\textwidth}
        \centering
        \includegraphics[width=\textwidth]{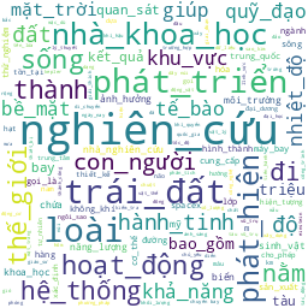}
        \caption{Science}
        \label{fig:sub8}
    \end{subfigure}
    \hfill
    \begin{subfigure}[b]{0.19\textwidth}
        \centering
        \includegraphics[width=\textwidth]{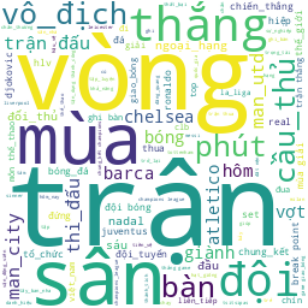}
        \caption{Sports}
        \label{fig:sub9}
    \end{subfigure}
    \hfill
    \begin{subfigure}[b]{0.19\textwidth}
        \centering
        \includegraphics[width=\textwidth]{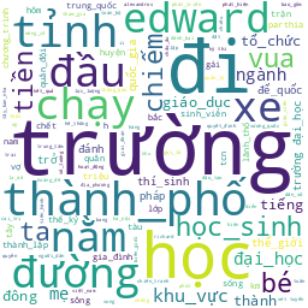}
        \caption{Other}
        \label{fig:sub10}
    \end{subfigure}

    \caption{Word clouds illustrating the most frequent words for each topic in the ViSP dataset.}
    \label{fig:main}
\end{figure*}

\subsubsection{Length Based Analysis}

Table~\ref{tab:sentence_length_distribution} shows the combined distribution of sentence lengths across both the original and paraphrased sentences. The majority of sentences are between 11 and 20 words, accounting for approximately 43.05\% of the dataset, which is the highest percentage among all length ranges. In contrast, sentences with more than 51 words represent the lowest percentage, comprising only about 0.65\% of the dataset.   

\begin{table}[ht]
\centering
\scalebox{0.92}{
\begin{tabular}{lcccc}
\toprule
\textbf{Words} & \textbf{Train} & \textbf{Validation} & \textbf{Test} & \textbf{All} \\ 
\midrule
\textbf{1-10}  & 10,111  & 4,009   & 3,983   & 18,103   \\
\textbf{11-20} & 56,043  & 23,800  & 23,657  & 103,500  \\
\textbf{21-30} & 42,945  & 18,737  & 18,419  & 80,101   \\
\textbf{31-40} & 16,291  & 6,414   & 6,255   & 28,960   \\
\textbf{41-50} & 4,975   & 1,514   & 1,651   & 8,140    \\
\textbf{51+}   & 826     & 355     & 372     & 1,553    \\
\bottomrule
\end{tabular}
}
\caption{Combined distribution of sentence lengths across the Train, Validation, Test, and All sets, including both original and paraphrase sentences.}
\label{tab:sentence_length_distribution}
\end{table}



\section{Experiments and Results}


\subsection{Human Performance}

Following human performance concept of other study like  \cite{vannguyen2020vietnamesedatasetevaluatingmachine, huynh-etal-2022-vinli}, we recruited five native Vietnamese speakers to perform the paraphrasing task. These individuals had no prior experience with paraphrasing tasks. Each annotator was asked to generate three paraphrases for a given set of sentences. Before starting, they were trained on the concept of paraphrasing and provided with guidelines to ensure that the paraphrases retained the original meaning while introducing lexical and structural variations.

Next, we randomly selected a subset of 300 samples, with 150 drawn from the test set and 150 from the validation set. This subset was designated as Test\textsubscript{300} for further evaluation.

\subsection{Re-Implemented Methods and Baselines}

In this section, we re-implemented the following method and models on our dataset as described in Section 3.

\textbf{EDA} \cite{wei2019eda} applies simple transformations such as random deletion (RD), random swap (RS), random insertion (RI), and synonym replacement (SR). For RI and SR, we replace WordNet with the PhoW2V model \cite{phow2v_vitext2sql} to generate Vietnamese synonyms.

\textbf{Back Translation} leverages translation between languages to produce semantically similar sentences. We use the en2vi and vi2en models from \cite{vinaitranslate} for this process.

We experiment with several pre-trained sequence-to-sequence models for paraphrase generation, including \textbf{mBART} \cite{tang2020multilingual}, \textbf{BARTpho} \cite{tran2021BARTpho}, \textbf{mT5} \cite{xue2020mt5}, and \textbf{ViT5} \cite{phan2022vit5}. These models were chosen for their strengths in both multilingual and Vietnamese-specific tasks. \textbf{mBART} and \textbf{mT5} provide robust multilingual capabilities, while \textbf{BARTpho} and \textbf{ViT5} are optimized for Vietnamese, offering language-specific nuances.

\subsection{Evaluation Metrics}

We use BLEU \cite{papineni2002bleu}, ROUGE \cite{lin-2004-rouge}, and BERTScore \cite{zhang2019bertscore} to evaluate paraphrase quality, and Distinct-N~\cite{li-etal-2016-diversity}, Entropy-N~\cite{shannon1948mathematical}, and Jaccard~\cite{jaccard1901etude} to measure diversity. For a detailed breakdown of evaluation metrics, see Appendix~\ref{append:metrics}.

\subsection{Experimental Settings}

We use a single NVIDIA Tesla A100 GPU via Google Colaboratory\footnote{\url{https://colab.research.google.com/}} to fine-tune all models on our dataset. When fine-tuning, we set the length of max length of sentence is 96 tokens, learning rate is 1e-5, batch size is 16 and training with five epochs. 


\subsection{Experimental Results}

\subsubsection{Single Paraphrase Evaluation}

\begin{table*}[h]
\centering
\scalebox{0.78}{
\begin{tabular}{lccccccccc}
\toprule
\textbf{Model}   & \multicolumn{3}{c}{\textbf{Val}} & \multicolumn{3}{c}{\textbf{Test}} & \multicolumn{3}{c}{\textbf{Test\textsubscript{300}}} \\
\cmidrule(lr){2-4} \cmidrule(lr){5-7} \cmidrule(lr){8-10}
                 & \small\textbf{BLEU-4} & \small\textbf{ROUGE-2} & \small\textbf{BERTScore} & \small\textbf{BLEU-4} & \small\textbf{ROUGE-2} & \small\textbf{BERTScore} & \small\textbf{BLEU-4} & \small\textbf{ROUGE-2} & \small\textbf{BERTScore} \\
\midrule
RD               & 26.82           & 52.49            & 66.19              & 26.87           & 52.42            & 66.10              & 26.23           & 52.93            & 66.27 \\
RS               & 14.79           & 48.89            & 63.75              & 14.76           & 48.70            & 63.76              & 14.69           & 48.86            & 64.12 \\
RI + PhoW2V      & 29.50           & 57.79            & 69.79              & 29.43           & 57.63            & 69.69              & 29.86           & 58.49            & 70.05 \\
SR + PhoW2V      & 22.81           & 46.48            & 63.30              & 22.67           & 46.46            & 63.28              & 22.86           & 46.14            & 63.10 \\
BT + vinai-translate-v2 & 54.33      & 63.84            & 79.23              & 53.97           & 63.65            & 21.60              & 54.03           & 64.33            & 79.99 \\
\midrule
mBART\textsubscript{large}            & 71.71           & 76.02            & \textbf{85.84}              & 71.12           & 75.76            & \textbf{85.74}              & \textbf{72.23}           & 76.20            & \textbf{86.17} \\
mT5\textsubscript{base}         & 60.22           & 70.20            & 81.46              & 59.58           & 69.69            & 81.27              & 60.84           & 71.00            & 81.85 \\
mT5\textsubscript{large}        & 27.04           & 46.59            & 72.02              & 26.86           & 46.23            & 71.84              & 26.83           & 46.48            & 72.38 \\
\midrule
BARTpho-syllable\textsubscript{base} & 68.39       & 73.97            & 84.34              & 67.66           & 73.51            & 84.15              & 70.39           & 75.03            & 85.06 \\
BARTpho-syllable\textsubscript{large} & 70.29           & 75.35            & 85.22              & 69.81           & 74.89            & 85.10              & 70.83           & 75.75            & 85.62 \\
BARTpho-word\textsubscript{base}& 69.61           & 74.63            & 79.25              & 68.76           & 74.18            & 79.15              & 70.23           & 75.47            & 79.72 \\
BARTpho-word\textsubscript{large}     & \textbf{72.06} & \textbf{76.06}  & 79.97    & \textbf{71.61} & \textbf{75.78}  & 79.99    & 71.70           & \textbf{76.22}  & 80.10 \\
ViT5\textsubscript{base}        & 70.20           & 74.91            & 85.08              & 69.75           & 74.58            & 85.00              & 71.24           & 75.69            & 85.37 \\
ViT5\textsubscript{large}       & 67.10           & 71.83            & 82.68              & 66.70           & 71.53            & 82.53              & 67.73           & 72.92            & 82.98 \\
\midrule
Human performance& -               & -                & -                  & -               & -                & -                  & 94.97           & 88.29            & 88.30 \\
\bottomrule
\end{tabular}
}
\caption{Evaluation of various models and methods on the Val, Test, and Test\textsubscript{300} sets of the ViSP dataset, assessing the best single paraphrased sentence generated by each model. The best overall results are highlighted in \textbf{bold}.}
\label{tab:exp1_one_to_one_evaluation}
\end{table*}

In the realm of single paraphrase generation, Table~\ref{tab:exp1_one_to_one_evaluation} shows that BARTpho-word\textsubscript{large} leads in BLEU-4 and ROUGE-2, with values of 72.06 and 76.06 on the Val set, respectively, indicating that the generated sentences are more similar to the reference paraphrases. It maintains strong performance on Test and Test\textsubscript{300}, achieving BLEU-4 of 71.61 and 71.70, and ROUGE-2 of 75.78 and 76.22. While mBART\textsubscript{large} performs well, it achieves higher BERTScore across all sets, with 85.84 on Val and 86.17 on Test\textsubscript{300}, suggesting that although the generated sentences differ more from the references, they retain better semantic similarity. ViT5-base also shows strong semantic preservation, with BERTScore reaching 85.37 on Test\textsubscript{300}. Among monolingual models, BARTpho-syllable\textsubscript{large} performs well, with BERTScore of 85.62 on Test\textsubscript{300}, reinforcing its effectiveness in generating faithful paraphrases. Human performance remains the upper bound, with BERTScore of 88.30, highlighting the gap between model-generated and human paraphrases. Among augmentation methods, Back Translation performs best, achieving a BERTScore of 79.99 on Test\textsubscript{300}, while simpler methods like Random Deletion and Synonym Replacement show notably lower scores.

\subsubsection{Multiple Paraphrases Evaluation}

\begin{table*}[ht]
\centering
\scalebox{0.78}{
\begin{tabular}{lccccccccc}
\toprule
\textbf{Model}   & \multicolumn{3}{c}{\textbf{K = 3}} & \multicolumn{3}{c}{\textbf{K = 5}} & \multicolumn{3}{c}{\textbf{K = 10}} \\
\cmidrule(lr){2-4} \cmidrule(lr){5-7} \cmidrule(lr){8-10}
                 & \small \textbf{BLEU-4} & \small \textbf{ROUGE-2} & \small \textbf{BERTScore} & \small\textbf{BLEU-4} & \small\textbf{ROUGE-2} & \small\textbf{BERTScore} & \small\textbf{BLEU-4} & \small\textbf{ROUGE-2} & \small\textbf{BERTScore} \\
\midrule
mBART\textsubscript{large}      & 70.35           & 75.23            & \textbf{85.29} & 69.76           & 74.84            & \textbf{85.06} & 66.16           & 71.50 & 81.57 \\
mT5\textsubscript{base}         & 58.68           & 68.85            & 80.32 & 57.78           & 68.27            & 79.88 & 54.68           & 66.36 & 78.36 \\
mT5\textsubscript{large}        & 26.98           & 46.39            & 71.51 & 27.07           & 46.48            & 71.46 & 27.29           & 46.70 & 71.09 \\
\midrule
BARTpho-syllable\textsubscript{base} & 66.81       & 72.99            & 83.63 & 66.20           & 72.56            & 83.37 & 62.26           & 68.94 & 79.54 \\
BARTpho-syllable\textsubscript{large} & 69.02           & 74.43            & 84.65 & 68.48           & 74.07            & 84.44 & 67.06           & 73.20 & \textbf{83.80} \\
BARTpho-word\textsubscript{base}& 68.03           & 73.70            & 78.50 & 67.42           & 73.31            & 78.34 & 64.09           & 70.21 & 75.54 \\
BARTpho-word\textsubscript{large}     & \textbf{70.88}           & \textbf{75.28}  & 79.39 & \textbf{70.31}           & \textbf{74.91}  & 79.23 & \textbf{69.10}           & \textbf{74.09} & 78.88 \\
ViT5\textsubscript{base}        & 68.93           & 74.08            & 84.52 & 68.35           & 73.73            & 84.34 & 66.92           & 72.81 & 83.68 \\
ViT5\textsubscript{large}       & 66.19           & 71.24            & 81.93 & 65.63           & 70.89            & 81.68 & 63.98           & 69.87 & 80.85 \\
\bottomrule
\end{tabular}
}
\caption{Evaluation of various models on the ViSP dataset's test set, focusing on K paraphrased sentences generated per input sentence (where K is the number of paraphrases). The BLEU-4, ROUGE-2 and BERTScore are averaged across all K paraphrases for each model. The best overall results are highlighted in \textbf{bold}. For a detailed breakdown of multiple paraphrase outputs, See Appendix~\ref{append:model_output}, Table~\ref{tab:model_raw_output}.}

\label{tab:exp2_one_to_many_evaluation}
\end{table*}

For multiple paraphrase generation, as shown in Table~\ref{tab:exp2_one_to_many_evaluation}, BARTpho-word\textsubscript{large} leads the performance across all K values, with the highest BLEU-4 and ROUGE-2 scores, indicating that its generated paraphrases closely resemble the reference sentences. mBART\textsubscript{large}, while slightly behind in BLEU-4 and ROUGE-2, achieves higher BERTScore for K = 3 and K = 5, suggesting strong semantic similarity. However, its performance declines at K = 10, showing reduced accuracy when generating more paraphrases. Among monolingual models with base architecture, ViT5\textsubscript{base} performs well at K = 10, achieving a BLEU-4 score of 66.92 and a BERTScore of 83.68. ViT5\textsubscript{base} also outperforms ViT5\textsubscript{large}, showing greater stability and less degradation in accuracy with increasing paraphrase numbers.

\subsubsection{Topic Based Evaluation}

\begin{figure*}[h]
    \centering
    \includegraphics[width=0.96 \textwidth]{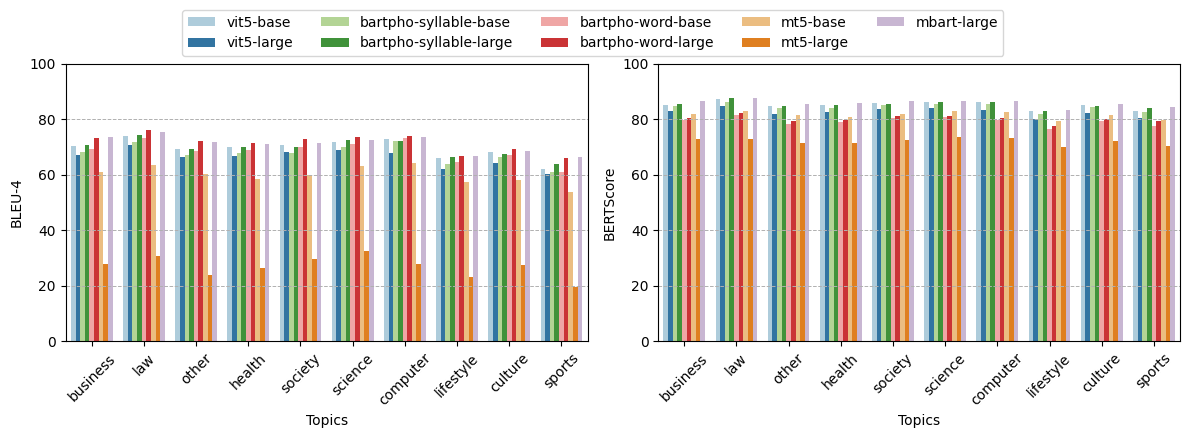}
    \caption{BLEU-4 and BERTScore across different topics.}
    \label{fig:bleu_rouge_topic}
\end{figure*}

As shown in Figure~\ref{fig:bleu_rouge_topic}, the results indicate that BARTpho-word\textsubscript{large} and mBART\textsubscript{large} consistently perform best across all topics. Within the T5 family, ViT5-base still outperforms other T5 models but also consistently surpasses all base BART models, including BARTpho-syllable\textsubscript{base} and BARTpho-word\textsubscript{base}. This is evident across multiple categories, especially in \textit{culture} and \textit{sports}. However, the challenge appears more pronounced when evaluated by BLEU, whereas BERTScore remains relatively similar across topics. Additionally, \textit{lifestyle}, \textit{culture}, and \textit{sports} emerge as the most challenging domains for all models, with the highest BLEU-4 scores in these categories hovering only around 70.

\subsubsection{Length Based Evaluation}

\begin{figure*}[h]
    \centering
    \includegraphics[width=0.96 \textwidth]{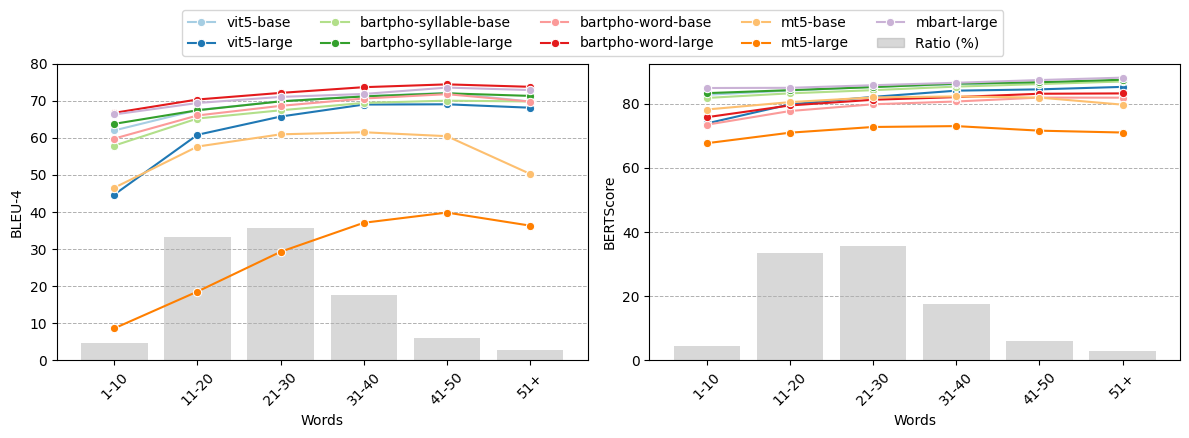}
    \caption{BLEU-4 and BERTScore across different lengths.}
    \label{fig:bleu_rouge_len}
\end{figure*}

As shown in Figure~\ref{fig:bleu_rouge_len}, BARTpho-word\textsubscript{large} consistently outperforms other models in BLEU across all sentence lengths, achieving the highest scores in the 41--50 word range. Meanwhile, BARTpho-syllable\textsubscript{large} tends to yield stronger BERTScore values, highlighting the overall effectiveness of BARTpho-based models. mBART\textsubscript{large} follows closely in most cases; for instance, in the 31--40 word range, BARTpho-syllable\textsubscript{base} achieves 69.38 BLEU-4 and 85.32 BERTScore, slightly below mBART. In contrast, T5-based models (ViT5, mT5) show weaker performance, particularly in shorter sentences (1--20 words). Notably, mT5\textsubscript{large} scores as low as 18.56 BLEU-4 and 70.96 BERTScore in the 11--20 word range, significantly below the BART-based models, which consistently perform well across all sentence lengths.

\subsubsection{Diversity Based Evaluation}

\begin{table*}[h]
\centering
\scalebox{0.78}{
\begin{tabular}{l cccc cccc cccc}
\toprule
\textbf{Model}   
& \multicolumn{4}{c}{\textbf{Val}}
& \multicolumn{4}{c}{\textbf{Test}}
& \multicolumn{4}{c}{\textbf{Test\textsubscript{300}}} \\
\cmidrule(lr){2-5} \cmidrule(lr){6-9} \cmidrule(lr){10-13}
& \small\textbf{DIST-1} & \small\textbf{DIST-2} & \small\textbf{ENT-4} & \small\textbf{Jaccard}
& \small\textbf{DIST-1} & \small\textbf{DIST-2} & \small\textbf{ENT-4} & \small\textbf{Jaccard}
& \small\textbf{DIST-1} & \small\textbf{DIST-2} & \small\textbf{ENT-4} & \small\textbf{Jaccard}\\
\midrule
mBART\textsubscript{large}
  & 95.67            & \underline{94.61} & 5.23  & 69.05
  & \underline{95.62} & \underline{94.59} & 5.22  & 68.82
  & 95.72            & \underline{94.60} & 4.34  & 70.31 \\
mT5\textsubscript{base}
  & 95.11            & 93.33            & 5.01  & 82.99
  & 94.91            & 93.27            & 5.01  & 82.72
  & 95.37            & 93.47            & 4.15  & 87.98 \\
mT5\textsubscript{large}
  & 56.30            & 58.94            & 5.02  & 72.25
  & 56.23            & 58.93            & 5.02  & 72.11
  & 54.62            & 56.77            & \textbf{4.51} & 72.75 \\
\midrule
BARTpho-syllable\textsubscript{base}
  & \underline{95.75} & 94.32            & 5.12  & 76.99
  & \underline{95.62} & 94.30            & 5.12  & 76.71
  & \textbf{96.17}    & 94.26            & 4.25  & 78.91 \\
BARTpho-syllable\textsubscript{large}
  & \textbf{95.90}  & 94.53            & 5.21  & 66.38
  & \textbf{95.75}  & 94.48            & 5.20  & 66.35
  & \underline{96.02} & 94.41            & 4.27  & 66.83 \\
BARTpho-word\textsubscript{base}
  & 93.95            & 94.60            & 5.21  & \underline{62.13}
  & 93.80            & 94.51            & 5.21  & \underline{62.10}
  & 94.09            & 94.36            & 4.41  & \underline{63.75} \\
BARTpho-word\textsubscript{large}
  & 93.85            & \textbf{94.77}  & \underline{5.27} & \textbf{58.00}
  & 93.72            & \textbf{94.70}  & \underline{5.28} & \textbf{57.99}
  & 93.82            & \textbf{94.63}  & \underline{4.47} & \textbf{58.26} \\
ViT5\textsubscript{base}
  & 95.33            & 94.15            & 5.14  & 75.70
  & 95.14            & 94.05            & 5.14  & 75.49
  & 95.48            & 94.01            & 4.27  & 77.36 \\
ViT5\textsubscript{large}
  & 93.99            & 93.59            & \textbf{5.30} & 68.78
  & 93.95            & 93.63            & \textbf{5.29} & 68.53
  & 94.75            & 94.04            & 4.39  & 69.89 \\
\midrule
Human performance
  & -            & -            & -  & -
  & -            & -            & -  & -
  & 95.54        & 94.96        & 6.48  & 52.33 \\
\bottomrule
\end{tabular}
}
\caption{Evaluation of the 5 beam-searched paraphrases in terms of DIST-1, DIST-2, ENT-4, and Jaccard on the Val, Test, and Test\textsubscript{300} sets. The \textbf{best} overall results are in bold, and the \underline{second best} are underlined.}
\label{tab:diversity_metrics_jaccard}
\end{table*}

Table~\ref{tab:diversity_metrics_jaccard} reports the diversity DIST-1, DIST-2 and entropy ENT-4 metrics, as well as the Jaccard scores, of the five beam-searched paraphrases generated by various models on the Test and Test\textsubscript{300} sets. Overall, BARTpho-word models exhibit strong performance, particularly BARTpho-word\textsubscript{large}, which achieves competitive bigram diversity with a DIST-2 score of 94.70 on the Test set and 94.63 on Test\textsubscript{300}, along with a high ENT-4 score of 5.28 on the Test set. Notably, BARTpho-syllable\textsubscript{large} attains the highest DIST-1 score of 95.75 on the Test set, while mBART\textsubscript{large} reaches the highest DIST-1 score of 95.72 on Test\textsubscript{300}. Meanwhile, ViT5\textsubscript{large} leads in ENT-4 for the Test set with a score of 5.29. Additionally, although mT5\textsubscript{base} achieves high Jaccard scores of 82.72 on the Test set and 87.98 on Test\textsubscript{300}, BARTpho-word\textsubscript{large} obtains the lowest Jaccard scores of 57.99 (Test) and 58.26 (Test\textsubscript{300}), signifying the greatest lexical diversity in its paraphrasing. These findings suggest that BARTpho-word excels at generating lexically varied and distributionally rich paraphrases, making it a robust option for applications requiring both diversity and consistency in paraphrasing.

\section{Discussion}

\textbf{Multilingual vs. Monolingual}. As \cite{conneau2019unsupervised} highlight, while multilingual models offer flexibility across languages, monolingual models often excel in specialized tasks due to their focus on a single language's nuances. This distinction becomes apparent in our test results, where BARTpho-word\textsubscript{large} demonstrates a clear advantage in both single and multi-paraphrase generation. Initially, BARTpho-word\textsubscript{large} holds a clear edge in single paraphrase generation, achieving the highest BLEU-4 and ROUGE-2 scores. Although mBART\textsubscript{large} achieves higher BERTScore in single paraphrase generation, it experiences a notable decline in performance as the number of required paraphrases increases (K=5, K=10), with BLEU-4 dropping to 66.16. This performance drop aligns with findings from \cite{hu2020xtreme}, which show that multilingual models struggle with generating numerous high-quality paraphrases. In contrast, monolingual models like BARTpho and ViT5 maintain strong performance in both single and multiple paraphrase tasks. Their focused training makes them better suited for tasks requiring high paraphrase diversity, consistently producing multiple outputs without losing quality.

\textbf{Impact of Model Architecture}. Research suggests that larger architectures do not always ensure better accuracy. For example, ViT5\textsubscript{large}, mT5\textsubscript{large} does not surpass ViT5\textsubscript{base}, mT5\textsubscript{base} accross all metrics, reflecting findings by \cite{kaplan2020scaling} that size increases do not guarantee performance gains. Particularly when generating less common words, larger models may perform worse in paraphrasing due to vocabulary deviations \cite{brown2020language}. In contrast, BART-based models, such as BARTpho-word and BARTpho-syllable, consistently show improvements in both accuracy and diversity with increased model size, as shown by \cite{lewis2019bart}, affirming the benefits of larger architectures in generating diverse paraphrases.

\textbf{Impact of Sentence Structure}. The structure of a sentence significantly influences the performance of paraphrase models. Different sentence types in Vietnamese—such as simple, compound, complex, and special—present varying levels of difficulty. As shown in Table~\ref{tab:sentence_structure_bleu_scores}, monolingual models like BARTpho-word\textsubscript{large} consistently achieve higher BLEU-4 scores for simple, compound, and complex sentences. This aligns with findings from \cite{isabelle2017challenge}, which suggest that models trained on a single language excel in capturing syntactic nuances. However, models often struggle with compound and complex sentences, which frequently include metaphorical and metonymic (See Appendix~\ref{append:dataset-detail}, Table~\ref{tab:visp_example_group_by_topic}) expressions in Vietnamese, as noted by \cite{shutova2013statistical}, highlighting the challenge of paraphrasing non-standard and figurative structures. These results suggest the need for improved handling of complex syntactic and figurative forms in paraphrasing tasks.


\begin{table}[h]
\centering
\scalebox{0.78}{
\begin{tabular}{lcccc}
\toprule
\textbf{Model} & \textbf{simple} & \textbf{compound} & \textbf{complex} \\
\midrule
mBART\textsubscript{large} & 73.06 & 67.63 & 69.73 \\
mT5\textsubscript{base} & 61.92 & 54.77 & 57.96 \\
mT5\textsubscript{large} & 24.80 & 27.23 & 31.14 \\
\midrule
BARTpho-syllable\textsubscript{base} & 69.58 & 64.24 & 66.27 \\
BARTpho-syllable\textsubscript{large} & 71.42 & 67.07 & 68.56 \\
BARTpho-word\textsubscript{base} & 70.44 & 65.45 & 67.69 \\
BARTpho-word\textsubscript{large} & \textbf{73.40} & \textbf{68.66} & \textbf{70.27} \\
ViT5\textsubscript{base} & 71.60 & 66.56 & 68.39 \\
ViT5\textsubscript{large} & 67.05 & 63.70 & 66.58 \\
\bottomrule
\end{tabular}
}
\caption{BLEU-4 scores of various models on different sentence structures in the ViSP Test set. The best overall results are highlighted in \textbf{bold}. For a detailed breakdown of sentence structs, see Appendix~\ref{append:dataset-detail}, Table~\ref{tab:visp_example_group_by_topic}.}
\label{tab:sentence_structure_bleu_scores}
\end{table}


\textbf{LLM Performance}. Table~\ref{tab:llm_eval} shows that Vietnamese-specific LLMs, such as Vistral-7B-Chat~\cite{van2023vistral}, lag behind significantly, indicating weaker paraphrase generation capabilities compared to general-purpose models. This suggests that current Vietnamese-focused models may require further optimization or fine-tuning for paraphrase tasks. Among multilingual and general-purpose LLMs, Meta-Llama-3.1-70B~\cite{dubey2024llama} achieves the best results, followed by Meta-Llama-3.1-8B~\cite{dubey2024llama}, GPT-4o~\cite{hurst2024gpt}, and Qwen2.5-7B~\cite{yang2024qwen2}, which all demonstrate strong lexical and semantic alignment with human paraphrases. However, none of the models reach human performance, indicating room for improvement in semantic fidelity and lexical variation. Since all models are evaluated without fine-tuning, their performance is reasonable, especially for high-resource models like Meta-Llama-3.1 and GPT-4o. However, fine-tuning on Vietnamese-specific paraphrase datasets could further narrow the gap between AI-generated and human paraphrases.

\begin{table}[h]
\centering
\scalebox{0.84}{
\begin{tabular}{lccc}
\toprule
\textbf{Model} & \small\textbf{BLEU-4} & \small\textbf{ROUGE-2} & \small\textbf{BERTScore} \\
\midrule
GPT-4o Mini              & 52.73  & 65.55  & 81.82  \\
Gemini 1.5 Flask                & 50.98  & 63.02  & 79.61  \\
\midrule
Vistral-7B-Chat            & 29.16  & 49.46  & 70.71  \\
Aya-23-8B                  & 42.15  & 59.52  & 75.21  \\
Qwen2.5-7B                 & 54.38  & 65.71  & 80.72  \\
Meta-Llama-3.1-8B          & 60.32  & 69.34  & 82.40  \\
Meta-Llama-3.1-70B         & \textbf{65.51}  & \textbf{73.21}  & \textbf{84.27}  \\
\midrule
Human Performance          & 94.97  & 88.29  & 88.30  \\
\bottomrule
\end{tabular}
}
\caption{Evaluation of various LLMs on the Test\textsubscript{300} using BLEU-4, ROUGE-2, and BERTScore metrics. The best overall results are highlighted in \textbf{bold}.}
\label{tab:llm_eval}
\end{table}

\section{Conclusion and Future Work}

We introduced ViSP, a Vietnamese paraphrase dataset created using human annotations and LLM outputs, for evaluating and benchmarking paraphrase generation models. We tested models like mBART, BARTpho, ViT5, and mT5 across various sentence lengths and topics, highlighting the strengths and weaknesses of multilingual and monolingual approaches. Our evaluation covered accuracy (BLEU, ROUGE, BERTScore) and diversity (Distinct-N, Entropy-N, Jaccard), and we compared model-generated paraphrases with human performance to assess the gap between automated systems and human paraphrasing.

In future, we plan to extend ViSP to tasks like machine translation, question answering and retrieval augmented generation. Additionally, we aim to pretrain a Vietnamese paraphrasing model, addressing a key gap in domain-specific models. This model will target complex linguistic phenomena, including metaphor and metonymy in Vietnamese, which present significant challenges for natural language understanding and generation tasks. ViSP will also support developing robust sentence similarity models like SBERT \cite{reimers2019sentence}, advancing Vietnamese NLP research.

\newpage

\section*{Limitations}

While our models were fine-tuned on the ViSP dataset, they were trained under low-resource conditions, which means the overall performance may not be fully optimized. With more computational resources, further improvements could be achieved. During data creation, we employed the Few-shot method to guide the generation process. However, we have not yet compared this approach with other advanced techniques like Chain-of-Thought \cite{wei2022chain} or Tree-of-Thought \cite{yao2024tree}, which could potentially yield better results in generating higher-quality paraphrases. Additionally, the current dataset lacks representation from certain specialized domains such as metaphor, mathematics and programing. This absence may affect the models ability to generalize to these specific areas.

\section*{Ethics Statement}
The ViSP dataset was developed with adherence to ethical guidelines. Human annotators were informed and compensated fairly. All datasets used, including UIT-ViQuAD \cite{vannguyen2020vietnamesedatasetevaluatingmachine}, UIT-ViNewsQA \cite{10.1145/3527631}, ALQAC \cite{nguyen2023summary} and ViNLI \cite{huynh-etal-2022-vinli}, were utilized in compliance with their respective licenses and terms of use. Additionally, in generating paraphrases with large language models (LLMs), we took steps to review and mitigate potential errors in the outputs, ensuring fairness and representativeness across different domains. 

\section*{Acknowledgement}

We sincerely appreciate the insightful comments and constructive feedback provided by the anonymous reviewers. This research is funded by Vietnam National University Ho Chi Minh City (VNU-HCM) under the grant number DS2025-26-01.


\bibliography{anthology,custom}

\begin{thebibliography}{51}
\expandafter\ifx\csname natexlab\endcsname\relax\def\natexlab#1{#1}\fi

\bibitem[{Alzantot et~al.(2018)Alzantot, Sharma, Elgohary, Ho, Srivastava, and Chang}]{alzantot2018generatingnaturallanguageadversarial}
Moustafa Alzantot, Yash Sharma, Ahmed Elgohary, Bo-Jhang Ho, Mani Srivastava, and Kai-Wei Chang. 2018.
\newblock \href {http://arxiv.org/abs/1804.07998} {Generating natural language adversarial examples}.

\bibitem[{Bernhard and Gurevych(2008)}]{bernhard2008answering}
Delphine Bernhard and Iryna Gurevych. 2008.
\newblock Answering learners’ questions by retrieving question paraphrases from social q\&a sites.

\bibitem[{Bhagat and Hovy(2013)}]{whatisparaphrase}
Rahul Bhagat and Eduard Hovy. 2013.
\newblock \href {https://doi.org/10.1162/COLI_a_00166} {{What Is a Paraphrase?}}
\newblock \emph{Computational Linguistics}, 39(3):463--472.

\bibitem[{Brown(2020)}]{brown2020language}
Tom~B Brown. 2020.
\newblock Language models are few-shot learners.
\newblock \emph{arXiv preprint arXiv:2005.14165}.

\bibitem[{Callison-Burch et~al.(2006)Callison-Burch, Koehn, and Osborne}]{callison2006improved}
Chris Callison-Burch, Philipp Koehn, and Miles Osborne. 2006.
\newblock Improved statistical machine translation using paraphrases.
\newblock In \emph{Proceedings of the Human Language Technology Conference of the NAACL, Main Conference}, pages 17--24.

\bibitem[{Conneau(2019)}]{conneau2019unsupervised}
A~Conneau. 2019.
\newblock Unsupervised cross-lingual representation learning at scale.
\newblock \emph{arXiv preprint arXiv:1911.02116}.

\bibitem[{Dong et~al.(2017)Dong, Mallinson, Reddy, and Lapata}]{dong2017learningparaphrasequestionanswering}
Li~Dong, Jonathan Mallinson, Siva Reddy, and Mirella Lapata. 2017.
\newblock \href {http://arxiv.org/abs/1708.06022} {Learning to paraphrase for question answering}.

\bibitem[{Dong et~al.(2021)Dong, Wan, and Cao}]{ParaSCI}
Qingxiu Dong, Xiaojun Wan, and Yue Cao. 2021.
\newblock \href {http://arxiv.org/abs/2101.08382} {Parasci: A large scientific paraphrase dataset for longer paraphrase generation}.

\bibitem[{Dubey et~al.(2024)Dubey, Jauhri, Pandey, Kadian, Al-Dahle, Letman, Mathur, Schelten, Yang, Fan et~al.}]{dubey2024llama}
Abhimanyu Dubey, Abhinav Jauhri, Abhinav Pandey, Abhishek Kadian, Ahmad Al-Dahle, Aiesha Letman, Akhil Mathur, Alan Schelten, Amy Yang, Angela Fan, et~al. 2024.
\newblock The llama 3 herd of models.
\newblock \emph{arXiv preprint arXiv:2407.21783}.

\bibitem[{Fleiss(1971)}]{fleiss1971measuring}
Joseph~L Fleiss. 1971.
\newblock Measuring nominal scale agreement among many raters.
\newblock \emph{Psychological bulletin}, 76(5):378.

\bibitem[{Gan and Ng(2019)}]{gan-ng-2019-improving}
Wee~Chung Gan and Hwee~Tou Ng. 2019.
\newblock \href {https://doi.org/10.18653/v1/P19-1610} {Improving the robustness of question answering systems to question paraphrasing}.
\newblock In \emph{Proceedings of the 57th Annual Meeting of the Association for Computational Linguistics}, pages 6065--6075, Florence, Italy. Association for Computational Linguistics.

\bibitem[{Grusky et~al.(2018)Grusky, Naaman, and Artzi}]{grusky2018newsroom}
Max Grusky, Mor Naaman, and Yoav Artzi. 2018.
\newblock Newsroom: A dataset of 1.3 million summaries with diverse extractive strategies.
\newblock \emph{arXiv preprint arXiv:1804.11283}.

\bibitem[{Hu et~al.(2020)Hu, Ruder, Siddhant, Neubig, Firat, and Johnson}]{hu2020xtreme}
Junjie Hu, Sebastian Ruder, Aditya Siddhant, Graham Neubig, Orhan Firat, and Melvin Johnson. 2020.
\newblock Xtreme: A massively multilingual multi-task benchmark for evaluating cross-lingual generalisation.
\newblock In \emph{International Conference on Machine Learning}, pages 4411--4421. PMLR.

\bibitem[{Hurst et~al.(2024)Hurst, Lerer, Goucher, Perelman, Ramesh, Clark, Ostrow, Welihinda, Hayes, Radford et~al.}]{hurst2024gpt}
Aaron Hurst, Adam Lerer, Adam~P Goucher, Adam Perelman, Aditya Ramesh, Aidan Clark, AJ~Ostrow, Akila Welihinda, Alan Hayes, Alec Radford, et~al. 2024.
\newblock Gpt-4o system card.
\newblock \emph{arXiv preprint arXiv:2410.21276}.

\bibitem[{Huynh et~al.(2022)Huynh, Nguyen, and Nguyen}]{huynh-etal-2022-vinli}
Tin~Van Huynh, Kiet~Van Nguyen, and Ngan Luu-Thuy Nguyen. 2022.
\newblock \href {https://aclanthology.org/2022.coling-1.339} {{V}i{NLI}: A {V}ietnamese corpus for studies on open-domain natural language inference}.
\newblock In \emph{Proceedings of the 29th International Conference on Computational Linguistics}, pages 3858--3872, Gyeongju, Republic of Korea. International Committee on Computational Linguistics.

\bibitem[{Isabelle et~al.(2017)Isabelle, Cherry, and Foster}]{isabelle2017challenge}
Pierre Isabelle, Colin Cherry, and George Foster. 2017.
\newblock A challenge set approach to evaluating machine translation.
\newblock \emph{arXiv preprint arXiv:1704.07431}.

\bibitem[{Jaccard(1901)}]{jaccard1901etude}
Paul Jaccard. 1901.
\newblock {\'E}tude comparative de la distribution florale dans une portion des alpes et des jura.
\newblock \emph{Bull Soc Vaudoise Sci Nat}, 37:547--579.

\bibitem[{Kaplan et~al.(2020)Kaplan, McCandlish, Henighan, Brown, Chess, Child, Gray, Radford, Wu, and Amodei}]{kaplan2020scaling}
Jared Kaplan, Sam McCandlish, Tom Henighan, Tom~B Brown, Benjamin Chess, Rewon Child, Scott Gray, Alec Radford, Jeffrey Wu, and Dario Amodei. 2020.
\newblock Scaling laws for neural language models.
\newblock \emph{arXiv preprint arXiv:2001.08361}.

\bibitem[{Landis(1977)}]{landis1977measurement}
JR~Landis. 1977.
\newblock The measurement of observer agreement for categorical data.
\newblock \emph{Biometrics}.

\bibitem[{Lewis(2019)}]{lewis2019bart}
M~Lewis. 2019.
\newblock Bart: Denoising sequence-to-sequence pre-training for natural language generation, translation, and comprehension.
\newblock \emph{arXiv preprint arXiv:1910.13461}.

\bibitem[{Li et~al.(2016)Li, Galley, Brockett, Gao, and Dolan}]{li-etal-2016-diversity}
Jiwei Li, Michel Galley, Chris Brockett, Jianfeng Gao, and Bill Dolan. 2016.
\newblock \href {https://doi.org/10.18653/v1/N16-1014} {A diversity-promoting objective function for neural conversation models}.
\newblock In \emph{Proceedings of the 2016 Conference of the North {A}merican Chapter of the Association for Computational Linguistics: Human Language Technologies}, pages 110--119, San Diego, California. Association for Computational Linguistics.

\bibitem[{Lin(2004)}]{lin-2004-rouge}
Chin-Yew Lin. 2004.
\newblock \href {https://aclanthology.org/W04-1013} {{ROUGE}: A package for automatic evaluation of summaries}.
\newblock In \emph{Text Summarization Branches Out}, pages 74--81, Barcelona, Spain. Association for Computational Linguistics.

\bibitem[{Lin et~al.(2014)Lin, Maire, Belongie, Hays, Perona, Ramanan, Doll{\'a}r, and Zitnick}]{mscoco}
Tsung-Yi Lin, Michael Maire, Serge Belongie, James Hays, Pietro Perona, Deva Ramanan, Piotr Doll{\'a}r, and C.~Lawrence Zitnick. 2014.
\newblock Microsoft coco: Common objects in context.
\newblock In \emph{Computer Vision -- ECCV 2014}, pages 740--755, Cham. Springer International Publishing.

\bibitem[{Long et~al.(2024)Long, Wang, Xiao, Zhao, Ding, Chen, and Wang}]{long2024llmsdrivensyntheticdatageneration}
Lin Long, Rui Wang, Ruixuan Xiao, Junbo Zhao, Xiao Ding, Gang Chen, and Haobo Wang. 2024.
\newblock \href {http://arxiv.org/abs/2406.15126} {On llms-driven synthetic data generation, curation, and evaluation: A survey}.

\bibitem[{Marceau et~al.(2022)Marceau, Belbahar, Queudot, Naji, Charton, and Meurs}]{marceau2022quickstartingdialogsystems}
Louis Marceau, Raouf Belbahar, Marc Queudot, Nada Naji, Eric Charton, and Marie-Jean Meurs. 2022.
\newblock \href {http://arxiv.org/abs/2204.02546} {Quick starting dialog systems with paraphrase generation}.

\bibitem[{Nguyen et~al.(2020{\natexlab{a}})Nguyen, Dao, and Nguyen}]{phow2v_vitext2sql}
Anh~Tuan Nguyen, Mai~Hoang Dao, and Dat~Quoc Nguyen. 2020{\natexlab{a}}.
\newblock {A Pilot Study of Text-to-SQL Semantic Parsing for Vietnamese}.
\newblock In \emph{Findings of the Association for Computational Linguistics: EMNLP 2020}, pages 4079--4085.

\bibitem[{Nguyen et~al.(2023{\natexlab{a}})Nguyen, Luu, Tran, Trieu, Dang, Nguyen, Nguyen, Pham, Pham, Vo et~al.}]{nguyen2023summary}
Chau Nguyen, Son~T Luu, Thanh Tran, An~Trieu, Anh Dang, Dat Nguyen, Hiep Nguyen, Tin Pham, Trang Pham, Thien-Trung Vo, et~al. 2023{\natexlab{a}}.
\newblock A summary of the alqac 2023 competition.
\newblock In \emph{2023 15th International Conference on Knowledge and Systems Engineering (KSE)}, pages 1--6. IEEE.

\bibitem[{Nguyen et~al.(2020{\natexlab{b}})Nguyen, Nguyen, Nguyen, and Nguyen}]{vannguyen2020vietnamesedatasetevaluatingmachine}
Kiet~Van Nguyen, Duc-Vu Nguyen, Anh Gia-Tuan Nguyen, and Ngan Luu-Thuy Nguyen. 2020{\natexlab{b}}.
\newblock \href {http://arxiv.org/abs/2009.14725} {A vietnamese dataset for evaluating machine reading comprehension}.

\bibitem[{Nguyen et~al.(2023{\natexlab{b}})Nguyen, Vo, Nguyen, Tran, and Van~Nguyen}]{nguyen2023viqp}
Sang~Quang Nguyen, Thuc~Dinh Vo, Duc~PA Nguyen, Dang~T Tran, and Kiet Van~Nguyen. 2023{\natexlab{b}}.
\newblock Viqp: Dataset for vietnamese question paraphrasing.
\newblock In \emph{2023 International Conference on Multimedia Analysis and Pattern Recognition (MAPR)}, pages 1--6. IEEE.

\bibitem[{Nguyen et~al.(2022)Nguyen, Nguyen, Phung, Nguyen, Tran, Luong, Vo, Bui, Phung, and Nguyen}]{vinaitranslate}
Thien~Hai Nguyen, Tuan-Duy~H. Nguyen, Duy Phung, Duy Tran-Cong Nguyen, Hieu~Minh Tran, Manh Luong, Tin~Duy Vo, Hung~Hai Bui, Dinh Phung, and Dat~Quoc Nguyen. 2022.
\newblock {A Vietnamese-English Neural Machine Translation System}.
\newblock In \emph{Proceedings of the 23rd Annual Conference of the International Speech Communication Association: Show and Tell (INTERSPEECH)}.

\bibitem[{Papineni et~al.(2002)Papineni, Roukos, Ward, and Zhu}]{papineni2002bleu}
Kishore Papineni, Salim Roukos, Todd Ward, and Wei-Jing Zhu. 2002.
\newblock Bleu: a method for automatic evaluation of machine translation.
\newblock In \emph{Proceedings of the 40th annual meeting of the Association for Computational Linguistics}, pages 311--318.

\bibitem[{Phan et~al.(2022)Phan, Tran, Nguyen, and Trinh}]{phan2022vit5}
Long Phan, Hieu Tran, Hieu Nguyen, and Trieu~H Trinh. 2022.
\newblock Vit5: Pretrained text-to-text transformer for vietnamese language generation.
\newblock \emph{arXiv preprint arXiv:2205.06457}.

\bibitem[{Reimers(2019)}]{reimers2019sentence}
N~Reimers. 2019.
\newblock Sentence-bert: Sentence embeddings using siamese bert-networks.
\newblock \emph{arXiv preprint arXiv:1908.10084}.

\bibitem[{Russo-Lassner et~al.(2005)Russo-Lassner, Lin, and Resnik}]{russo2005paraphrase}
Grazia Russo-Lassner, Jimmy Lin, and Philip Resnik. 2005.
\newblock A paraphrase-based approach to machine translation evaluation.
\newblock \emph{LAMP-TR-125 CS-TR-4754 UMIACS-TR-2005-57, University of Maryland, College Park, MD}.

\bibitem[{Scherrer(2020)}]{scherrer-2020-tapaco}
Yves Scherrer. 2020.
\newblock \href {https://aclanthology.org/2020.lrec-1.848} {{T}a{P}a{C}o: A corpus of sentential paraphrases for 73 languages}.
\newblock In \emph{Proceedings of the Twelfth Language Resources and Evaluation Conference}, pages 6868--6873, Marseille, France. European Language Resources Association.

\bibitem[{Shannon(1948)}]{shannon1948mathematical}
Claude~Elwood Shannon. 1948.
\newblock A mathematical theory of communication.
\newblock \emph{The Bell system technical journal}, 27(3):379--423.

\bibitem[{Shutova et~al.(2013)Shutova, Teufel, and Korhonen}]{shutova2013statistical}
Ekaterina Shutova, Simone Teufel, and Anna Korhonen. 2013.
\newblock Statistical metaphor processing.
\newblock \emph{Computational Linguistics}, 39(2):301--353.

\bibitem[{Tang et~al.(2020)Tang, Tran, Li, Chen, Goyal, Chaudhary, Gu, and Fan}]{tang2020multilingual}
Yuqing Tang, Chau Tran, Xian Li, Peng-Jen Chen, Naman Goyal, Vishrav Chaudhary, Jiatao Gu, and Angela Fan. 2020.
\newblock Multilingual translation with extensible multilingual pretraining and finetuning.
\newblock \emph{arXiv preprint arXiv:2008.00401}.

\bibitem[{Team et~al.(2023)Team, Anil, Borgeaud, Wu, Alayrac, Yu, Soricut, Schalkwyk, Dai, Hauth et~al.}]{team2023gemini}
Gemini Team, Rohan Anil, Sebastian Borgeaud, Yonghui Wu, Jean-Baptiste Alayrac, Jiahui Yu, Radu Soricut, Johan Schalkwyk, Andrew~M Dai, Anja Hauth, et~al. 2023.
\newblock Gemini: a family of highly capable multimodal models.
\newblock \emph{arXiv preprint arXiv:2312.11805}.

\bibitem[{Tran et~al.(2021)Tran, Le, and Nguyen}]{tran2021BARTpho}
Nguyen~Luong Tran, Duong~Minh Le, and Dat~Quoc Nguyen. 2021.
\newblock Bartpho: pre-trained sequence-to-sequence models for vietnamese.
\newblock \emph{arXiv preprint arXiv:2109.09701}.

\bibitem[{Van~Nguyen et~al.(2023)Van~Nguyen, Nguyen, Nguyen, Nguyen, Pl{\"u}ster, Pham, Nguyen, Schramowski, and Nguyen}]{van2023vistral}
Chien Van~Nguyen, Thuat Nguyen, Quan Nguyen, Huy Nguyen, Bj{\"o}rn Pl{\"u}ster, Nam Pham, Huu Nguyen, Patrick Schramowski, and Thien Nguyen. 2023.
\newblock Vistral-7b-chat-towards a state-of-the-art large language model for vietnamese.

\bibitem[{Van~Nguyen et~al.(2022)Van~Nguyen, Van~Huynh, Nguyen, Nguyen, and Nguyen}]{10.1145/3527631}
Kiet Van~Nguyen, Tin Van~Huynh, Duc-Vu Nguyen, Anh Gia-Tuan Nguyen, and Ngan Luu-Thuy Nguyen. 2022.
\newblock \href {https://doi.org/10.1145/3527631} {New vietnamese corpus for machine reading comprehension of health news articles}.
\newblock \emph{ACM Trans. Asian Low-Resour. Lang. Inf. Process.}, 21(5).

\bibitem[{Wallis(1993)}]{wallis1993information}
Peter Wallis. 1993.
\newblock Information retrieval based on paraphrase.
\newblock In \emph{Proceedings of pacling conference}. Citeseer.

\bibitem[{Wei et~al.(2022)Wei, Wang, Schuurmans, Bosma, Xia, Chi, Le, Zhou et~al.}]{wei2022chain}
Jason Wei, Xuezhi Wang, Dale Schuurmans, Maarten Bosma, Fei Xia, Ed~Chi, Quoc~V Le, Denny Zhou, et~al. 2022.
\newblock Chain-of-thought prompting elicits reasoning in large language models.
\newblock \emph{Advances in neural information processing systems}, 35:24824--24837.

\bibitem[{Wei and Zou(2019)}]{wei2019eda}
Jason Wei and Kai Zou. 2019.
\newblock Eda: Easy data augmentation techniques for boosting performance on text classification tasks.
\newblock \emph{arXiv preprint arXiv:1901.11196}.

\bibitem[{Xue(2020)}]{xue2020mt5}
L~Xue. 2020.
\newblock mt5: A massively multilingual pre-trained text-to-text transformer.
\newblock \emph{arXiv preprint arXiv:2010.11934}.

\bibitem[{Yang et~al.(2024)Yang, Yang, Zhang, Hui, Zheng, Yu, Li, Liu, Huang, Wei et~al.}]{yang2024qwen2}
An~Yang, Baosong Yang, Beichen Zhang, Binyuan Hui, Bo~Zheng, Bowen Yu, Chengyuan Li, Dayiheng Liu, Fei Huang, Haoran Wei, et~al. 2024.
\newblock Qwen2. 5 technical report.
\newblock \emph{arXiv preprint arXiv:2412.15115}.

\bibitem[{Yao et~al.(2024)Yao, Yu, Zhao, Shafran, Griffiths, Cao, and Narasimhan}]{yao2024tree}
Shunyu Yao, Dian Yu, Jeffrey Zhao, Izhak Shafran, Tom Griffiths, Yuan Cao, and Karthik Narasimhan. 2024.
\newblock Tree of thoughts: Deliberate problem solving with large language models.
\newblock \emph{Advances in Neural Information Processing Systems}, 36.

\bibitem[{Zhang et~al.(2019)Zhang, Kishore, Wu, Weinberger, and Artzi}]{zhang2019bertscore}
Tianyi Zhang, Varsha Kishore, Felix Wu, Kilian~Q Weinberger, and Yoav Artzi. 2019.
\newblock Bertscore: Evaluating text generation with bert.
\newblock \emph{arXiv preprint arXiv:1904.09675}.

\bibitem[{Zhu et~al.(2023)Zhu, Zhang, Haq, Hui, and Tyson}]{zhu2023chatgptreproducehumangeneratedlabels}
Yiming Zhu, Peixian Zhang, Ehsan-Ul Haq, Pan Hui, and Gareth Tyson. 2023.
\newblock \href {http://arxiv.org/abs/2304.10145} {Can chatgpt reproduce human-generated labels? a study of social computing tasks}.

\bibitem[{Zukerman et~al.(2002)Zukerman, Raskutti, and Wen}]{zukerman2002experiments}
Ingrid Zukerman, Bhavani Raskutti, and Yingying Wen. 2002.
\newblock Experiments in query paraphrasing for information retrieval.
\newblock In \emph{Australian Joint Conference on Artificial Intelligence}, pages 24--35. Springer.

\end{thebibliography}
\bibliographystyle{acl_natbib}

\newpage
\appendix

\onecolumn
\section*{Appendix}
\renewcommand{\thesubsection}{\Alph{subsection}}
In this section, we provide supplementary information to support the main content of this paper. This includes additional details about the datasets, models, evaluation metrics, and methods used throughout our experiments. 
\subsection{Dataset Details}
\label{append:dataset-detail}
The ViSP dataset is compiled from several publicly available sources, including:

\begin{enumerate}
 \item\textbf{UIT-ViQuAD} \cite{vannguyen2020vietnamesedatasetevaluatingmachine} This machine reading comprehension dataset includes over 23,000 human-generated question-answer pairs. These pairs are derived from 5,109 passages extracted from 174 Vietnamese Wikipedia articles, providing a rich source of information and ensuring a diverse range of topics and contexts. 
 
 \item\textbf{UIT-ViNewsQA} \cite{10.1145/3527631} This corpus consists of 22,057 question-answer pairs created by crowd-workers. These pairs are based on a collection of 4,416 Vietnamese healthcare news articles, with answers being textual spans directly taken from the corresponding articles. 
 
\item\textbf{ALQAC} \cite{nguyen2023summary}: The ALQAC dataset contains thousands of multiple-choice question-answer pairs, sourced from Vietnamese legal documents. Each pair is carefully reviewed for clarity and accuracy, making it an essential resource for testing question answering models in the legal domain.
 
\item\textbf{ViNLI} \cite{huynh-etal-2022-vinli} The ViNLI corpus comprises over 30,000 human-annotated premise-hypothesis sentence pairs. These pairs are extracted from more than 800 online news articles, offering a substantial and varied dataset for natural language inference tasks. 

\end{enumerate}

\begin{figure*}[h]
    \centering
    \includegraphics[width=0.8 \textwidth]{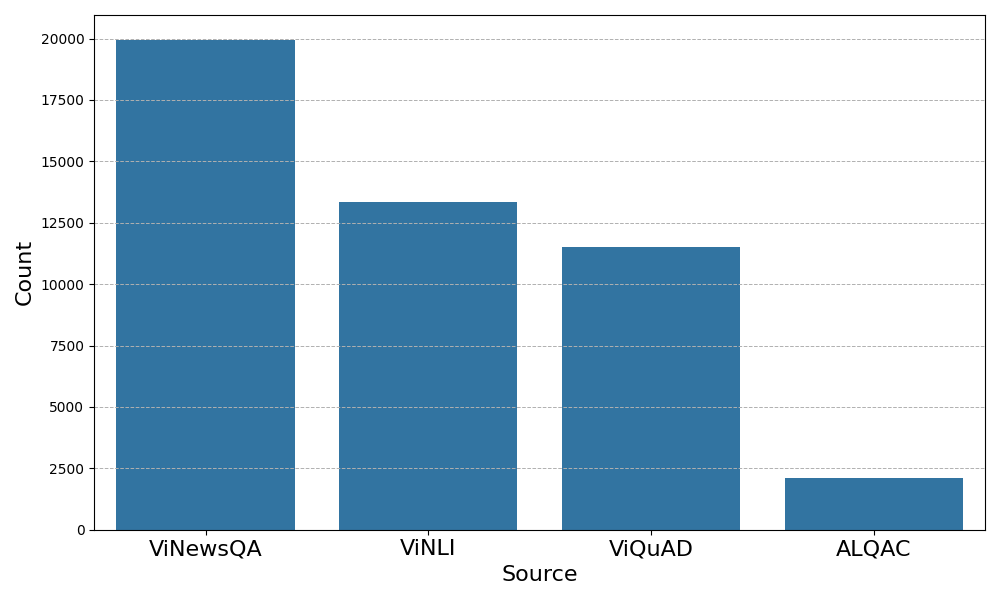}
    \caption{Distribution of sentence source in the ViSP dataset.}
    \label{fig:source_dis}
\end{figure*}

Figure~\ref{fig:source_dis} shows the distribution of original sentence sources in the ViSP dataset, with UIT-ViNewsQA and ViNLI contributing the largest proportion. This suggests that ViSP is heavily influenced by news-related content, which may impact the linguistic patterns and domain coverage of the paraphrases.

\begin{table*}[h]
\centering
\begin{tabular}{p{10cm}p{1.5cm}p{1cm}p{1.5cm}}
\toprule
\textbf{Sentence} & \textbf{Source} & \textbf{Topic} & \textbf{Structure} \\
\midrule
\small Vào ngày 23 tháng 9 năm 1846, nhà thiên văn Johann Galle đã phát hiện ra Sao Hải Vương ở vị trí lệch 1 độ so với tiên đoán của Urbain Le Verrier. (\textit{English: On September 23, 1846, astronomer Johann Galle discovered Neptune 1 degree off from Urbain Le Verrier's prediction.})& \small ViQuAD & \small Science & \small Complex \\
\midrule
\small Theo giáo sư Long, mỗi nước có khuyến cáo khác nhau khi điều trị vi khuẩn HP. (\textit{English: According to Professor Long, each country has different recommendations when treating HP bacteria.})& \small ViNewsQA & \small Health & \small Simple \\
\midrule
\small \textcolor{red}{Hazard gia nhập Real} hè 2019 từ \textcolor{red}{Chelsea}, theo bản hợp đồng trị giá 190 triệu USD - trong đó có 112 triệu USD trả trước. (\textit{English: \textcolor{red}{Hazard joined Real} in the summer of 2019 from \textcolor{red}{Chelsea}, in a contract worth 190 million USD - including 112 million USD in advance.}) \textbf{\textcolor{red}{\#metonym}} & \small ViNLI & \small Sports & \small Compound \\
\midrule
\small Công ty đã hứng rất nhiều cuộc tấn công mạng, phải từ bỏ nhiều dịch vụ chủ chốt trước khi \textcolor{blue}{'bán mình'} cho \textcolor{red}{đại gia viễn thông Mỹ Verizon}. (\textit{English: The company faced numerous cyberattacks and had to abandon several key services before \textcolor{blue}{'selling itself'} to \textcolor{red}{the American telecom giant Verizon.}}) \textbf{\textcolor{red}{\#metonym}} \textbf{\textcolor{blue}{\#metaphor}} & \small ViNLI & \small Business & \small Compound \\
\midrule
\small Trong khi đó, Điện Kremlin tuyên bố đang nghiên cứu khả năng tổ chức hội nghị này. (\textit{English: Meanwhile, \textcolor{red}{the Kremlin} announced that it is studying the possibility of holding this conference.}) \textbf{\textcolor{red}{\#metonym}} & \small ViQuAD & \small Society & \small Single \\
\midrule
\small Những người có làn da ngăm đen tạo cảm giác khỏe mạnh, gợi cảm cho người đối diện. (\textit{English: People with dark skin give a feeling of health and sexiness to the other person.})& \small ViNewsQA & \small Lifestyle & \small Simple \\
\midrule
\small Đơn khiếu nại phải kèm theo bản sao quyết định giải quyết khiếu nại lần đầu và các tài liệu kèm theo. (\textit{English: The complaint must be accompanied by a copy of the initial complaint resolution decision and accompanying documents.})& \small ALQAC & \small Law & \small Complex \\
\midrule
\small Ellen DeGeneres, sinh năm 1958, là ngôi sao truyền hình hàng đầu tại Mỹ. (\textit{English: Ellen DeGeneres, born in 1958, is a top television star in America.})& \small ViNLI & \small Culture & \small Simple \\
\midrule
\small Video đầu tiên hiện có hơn 2,1 triệu lượt xem chỉ sau một ngày đăng tải. (\textit{English: The first video now has more than 2.1 million views after just one day of posting.})& \small ViNLI & \small Other & \small Simple \\
\bottomrule
\end{tabular}
\caption{Examples of classifying sentences by topic in ViSP dataset.}
\label{tab:visp_example_group_by_topic}
\end{table*}

\newpage
\subsection{Metrics}
\label{append:metrics}


\subsubsection{Sematic and Diversity}
\begin{enumerate}

    \item \textbf{BLEU-4}~\cite{papineni2002bleu}: This metric measures the precision of 4-grams between the generated paraphrase and the reference. A higher BLEU-4 value indicates greater syntactic and lexical alignment with the reference.
    
    \item \textbf{ROUGE-2}~\cite{lin-2004-rouge}: This metric calculates the recall of bigrams (2-grams) in the generated paraphrase compared to the reference. A higher ROUGE-2 value reflects better preservation of key content from the reference.
    
    \item \textbf{BERTScore}~\cite{zhang2019bertscore}: This measure uses contextual embeddings to compare each token in the generated paraphrase with those in the reference. A higher BERTScore implies stronger semantic similarity and fidelity to the reference text.
    
    \item \textbf{DIST-1 and DIST-2}~\cite{li-etal-2016-diversity}: These metrics capture the distinctiveness of unigrams and bigrams, respectively. Higher values indicate more diverse and less repetitive paraphrases.
    
    \item \textbf{ENT-4}~\cite{shannon1948mathematical}: This is the entropy of 4-grams, reflecting the diversity and unpredictability of word combinations. A higher ENT-4 value suggests more varied and creative paraphrases.
    
    \item \textbf{Jaccard}~\cite{jaccard1901etude}: This score measures the lexical overlap between the original sentence and its paraphrase. A lower Jaccard value indicates less overlap with the source, and hence greater paraphrase diversity.

\end{enumerate}

\subsubsection{Human Eval}
\label{append:human-eval}
We conduct manual evaluations where human reviewers assess the quality of paraphrased sentences. Each paraphrase is evaluated based on four key criteria, with reviewers assigning a score from 1 (poor) to 5 (excellent) for each criterion:

\begin{enumerate}
    \item \textbf{INF (Informativeness)}: How well does the paraphrase retain the original meaning?    
    \item \textbf{REL (Relevance)}: To what extent are the important facts and details preserved?
    \item \textbf{FLU (Fluency)}: How fluent and natural does the sentence sound?
    \item \textbf{COH (Coherence)}: How well do the sentence parts fit together to form a coherent whole?
\end{enumerate}

\subsection{Paraphrase Verification Checklist}
\label{append:paraphrase-checklist}

To ensure the quality and accuracy of the paraphrased sentences in our dataset, we implemented a verification process where annotators assessed whether each sentence pair constituted a valid paraphrase. Annotators were instructed to evaluate each sentence pair based on the above criteria in Table~\ref{tab:paraphrase_checklist}. If the paraphrased sentence met all the criteria, it was marked as a valid paraphrase. If it failed to meet any of the criteria, it was marked as invalid.

\begin{table*}[h]
\centering
\begin{tabular}{p{0.3\linewidth} p{0.65\linewidth}}
\toprule
\textbf{Rule} & \textbf{Question} \\
\midrule
    \small\textbf{\MakeUppercase{Semantic Equivalence}} & \small Does the paraphrased sentence convey the same meaning as the original sentence, preserving all key information without adding or omitting important details? \\
    \small\textbf{\MakeUppercase{Fluency \& Grammatical}} & \small Is the paraphrased sentence grammatically correct and fluent in Vietnamese? \\
    \small\textbf{\MakeUppercase{Style \& Tone Consistency}} & \small Does the paraphrase maintain the same style and tone as the original sentence? \\
    \small\textbf{\MakeUppercase{No Contradictions}} & \small Does the paraphrase avoid contradicting any facts or statements in the original sentence? \\
\bottomrule
\end{tabular}
\caption{Checklist used by annotators to verify if a sentence pair is a valid paraphrase. 
}
\label{tab:paraphrase_checklist}
\end{table*}

\begin{figure}[h]
    \centering
    \includegraphics[width=1 \textwidth]{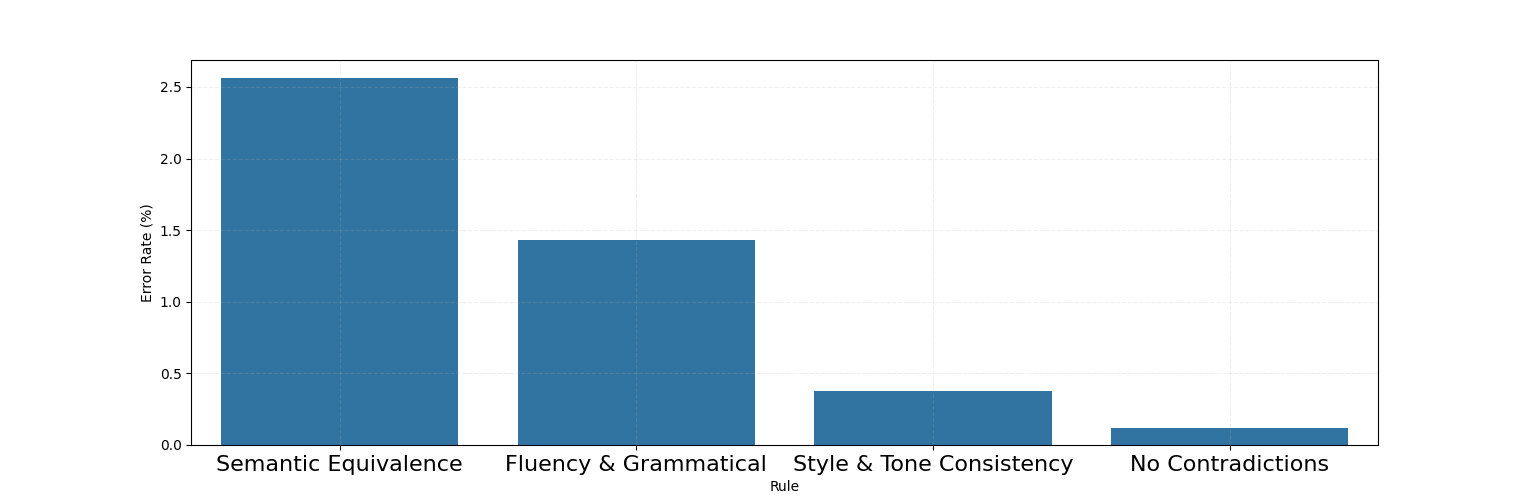}
    \caption{Error rate distribution across different paraphrase verification rule.}
    \label{fig:error_dist}
\end{figure}

Figure~\ref{fig:error_dist} presents the distribution of errors across different paraphrase verification rules. The analysis reveals that \MakeUppercase{Semantic Equivalence} is the most common source of errors, indicating that paraphrased sentences often fail to fully preserve the meaning of the original text. This suggests that maintaining semantic consistency remains a significant challenge in paraphrase generation. Additionally, while \MakeUppercase{No Contradictions} and \MakeUppercase{Style \& Tone Consistency} exhibit lower error rates, \MakeUppercase{Fluency \& Grammatical} still accounts for a noticeable portion of errors. 

\subsection{Model Outputs}
\label{append:model_output}
In Table~\ref{tab:model_raw_output}, we present example outputs from the models across various experiments to further clarify and support the findings discussed in the main text. These supplementary results include detailed paraphrases generated by models like mBART, BARTpho, ViT5, and mT5. 


\begin{table*}[h]
\centering
\begin{tabular}{p{0.15\linewidth} p{0.8\linewidth}}
\toprule 
\textbf{Model} & \textbf{Paraphrases} \\
\midrule

Ground Truth &
\small \textcolor{blue}{Đại lễ săn thỏ Phục sinh năm nay} là lần đầu tiên được tổ chức \textcolor{blue}{kể từ năm} 2017. (\textit{English: \textcolor{blue}{This year's Easter Bunny Hunt} is the first to be held \textcolor{blue}{since 2017.}})

\small \textcolor{blue}{Sau 2017,} năm nay là lần đầu tiên Đại lễ săn thỏ Phục sinh được tổ chức. (\textit{English: \textcolor{blue}{After 2017}, this year is the first time the Easter Bunny Hunt is held.})

\small Đại lễ săn thỏ Phục sinh đã được tổ chức \textcolor{blue}{trở lại sau 5 năm vắng bóng,} kể từ năm 2017. (\textit{English: The Easter Bunny Hunt has been held \textcolor{blue}{again after a 5-year absence}, since 2017.})


\small \textcolor{blue}{Từ năm 2017 đến nay,} Đại lễ săn thỏ Phục sinh đã không được tổ chức, \textcolor{blue}{cho đến năm nay.} (\textit{English: \textcolor{blue}{Since 2017,} the Easter Bunny Hunt has not been held, \textcolor{blue}{until this year.}})
\\
\midrule
mBART\textsubscript{large} &
\small Đại lễ săn thỏ Phục sinh \textcolor{blue}{đã được tổ chức lần đầu tiên} kể từ \textcolor{blue}{năm} 2017. 

\small Đại lễ săn thỏ Phục sinh \textcolor{blue}{được tổ chức lần đầu tiên} kể từ \textcolor{blue}{năm} 2017. 

\small Đại lễ săn thỏ Phục sinh \textcolor{blue}{năm nay lần đầu tiên} được tổ chức kể từ \textcolor{blue}{năm} 2017. 
\\
\midrule
mT5\textsubscript{base} &
\small \textcolor{blue}{Năm nay,} Đại lễ săn thỏ Phục sinh được tổ chức kể từ 2017. 

\small \textcolor{blue}{Vào năm nay,} Đại lễ săn thỏ Phục sinh được tổ chức kể từ 2017.

\small \textcolor{blue}{Năm nay,} Đại lễ săn thỏ Phục sinh \textcolor{blue}{đã} được tổ chức kể từ 2017.
\\
\midrule
mT5\textsubscript{large} &
\small \textcolor{blue}{Săn thỏ Phục sinh} được tổ chức kể từ 2017.

\small \textcolor{blue}{Này} là lần đầu tiên Đại lễ săn thỏ Phục sinh được tổ chức kể từ 2017. 

\small \textcolor{blue}{Săn thỏ Phục sinh} được tổ chức kể từ 2017 \textcolor{blue}{năm nay}. 
\\
\midrule
BARTpho-syllable\textsubscript{base} &
\small \textcolor{blue}{Đây} là lần đầu tiên Đại lễ săn thỏ Phục sinh được tổ chức kể từ \textcolor{blue}{năm} 2017. 

\small \textcolor{blue}{Kể từ năm 2017}, Đại lễ săn thỏ Phục sinh đã được tổ chức lần đầu tiên .  

\small Đại lễ săn thỏ Phục sinh \textcolor{blue}{năm nay} là lần đầu tiên được tổ chức kể từ 2017. 
\\
\midrule
BARTpho-syllable\textsubscript{large} &
\small Đại lễ săn thỏ Phục sinh \textcolor{blue}{năm nay đánh dấu} lần đầu tiên \textcolor{blue}{Đại lễ} được tổ chức kể từ \textcolor{blue}{năm} 2017. 

\small Đại lễ săn thỏ Phục sinh \textcolor{blue}{năm nay đánh dấu} lần đầu tiên được tổ chức kể từ \textcolor{blue}{năm} 2017. 

\small Đại lễ săn thỏ Phục sinh \textcolor{blue}{năm nay đánh dấu} lần đầu tiên \textcolor{blue}{Đại lễ này} được tổ chức kể từ \textcolor{blue}{năm} 2017. 
\\
\midrule
BARTpho-word\textsubscript{base} &
\small Đại lễ săn thỏ Phục sinh \textcolor{blue}{đã được tổ chức lần đầu tiên} kể từ \textcolor{blue}{năm} 2017. 

\small Đại lễ săn thỏ Phục sinh \textcolor{blue}{năm nay} là lần đầu tiên được tổ chức kể từ 2017. 

\small Đại lễ săn thỏ Phục sinh \textcolor{blue}{đã được tổ chức thường niên} kể từ \textcolor{blue}{năm} 2017. 
\\
\midrule
BARTpho-word\textsubscript{large} &
\small Đại lễ săn thỏ Phục sinh \textcolor{blue}{năm nay là lần đầu tiên sự kiện này} được tổ chức kể từ \textcolor{blue}{năm} 2017. 

\small Đại lễ săn thỏ Phục sinh \textcolor{blue}{năm nay đánh dấu lần đầu tiên lễ hội này} được tổ chức kể từ \textcolor{blue}{năm} 2017. 

\small Đại lễ săn thỏ Phục sinh \textcolor{blue}{năm nay là sự kiện đầu tiên} được tổ chức kể từ \textcolor{blue}{năm} 2017.  
\\
\midrule
ViT5\textsubscript{base} &
\small Đại lễ săn thỏ Phục sinh \textcolor{blue}{năm nay} là lần đầu tiên được tổ chức kể từ \textcolor{blue}{năm} 2017. 

\small Đại lễ săn thỏ Phục sinh \textcolor{blue}{năm nay lần đầu tiên} được tổ chức kể từ \textcolor{blue}{năm} 2017. 

\small Đại lễ săn thỏ Phục sinh \textcolor{blue}{năm nay là lần đầu tiên được tổ chức tại Việt Nam} kể từ \textcolor{blue}{năm} 2017.
\\
\midrule
ViT5\textsubscript{large} &
\small \textcolor{blue}{Việc tổ chức} Đại lễ săn thỏ Phục sinh \textcolor{blue}{năm nay} là lần đầu tiên được tổ chức kể từ \textcolor{blue}{năm} 2017. 

\small \textcolor{blue}{Việc tổ chức} Đại lễ săn thỏ Phục sinh là lần đầu tiên được tổ chức kể từ \textcolor{blue}{năm} 2017. 
 
\small \textcolor{blue}{Người ta đã tổ chức} Đại lễ săn thỏ Phục sinh lần đầu tiên kể từ \textcolor{blue}{năm} 2017. 
\\
\bottomrule 
\end{tabular}

\caption{Examples of paraphrases generated by various models for the original sentence "Năm nay là lần đầu tiên Đại lễ săn thỏ Phục sinh được tổ chức kể từ 2017." (\textit{English: This year is the first time the Easter Bunny Hunt has been held since 2017.}). The highlighted show the lexical differences compared to the original sentence.}
\label{tab:model_raw_output}
\end{table*}

\end{document}